\documentclass[preprints,article,accept,moreauthors,pdftex]{mdpi} 

\firstpage{1} 
\pubvolume{xx}
\issuenum{1}
\articlenumber{5}
\pubyear{2019}
\copyrightyear{2019}
\history{Received: date; Accepted: date; Published: date}

\pdfoutput=1

\usepackage{algorithm}
\usepackage{algorithmic}
\usepackage{bbm}
\usepackage{amssymb}
\usepackage{chngcntr}

\makeatletter
\newcommand{\dontusepackage}[2][]{%
  \@namedef{ver@#2.sty}{9999/12/31}%
  \@namedef{opt@#2.sty}{#1}}
\makeatother
\dontusepackage[bar,baz]{hyperref}

\Title{Non-linear Canonical Correlation Analysis:\\ A Compressed Representation Approach}

\Author{Amichai Painsky  $^{1,}$*, Meir Feder  $^{2}$ and Naftali Tishby $^{3}$}

\AuthorNames{Firstname Lastname, Firstname Lastname and Firstname Lastname}

\address{%
$^{1}$ \quad The Industrial Engineering Department, Tel Aviv University, Israel, 6997801; amichaip@tauex.tau.ac.il\\
$^{2}$ \quad The School of Electrical Engineering, Tel Aviv University, Israel, 6997801; meir@tauex.tau.ac.il\\
$^{3}$ \quad The School of Computer Science and Engineering and 
       the Interdisciplinary Center for Neural Computation, The Hebrew University of Jerusalem, Israel, 9190401; tishby@cs.huji.ac.il}

\corres{Correspondence: amichaip@tauex.tau.ac.il}

\abstract{Canonical Correlation Analysis (CCA) is a linear representation learning method that seeks maximally correlated variables in multi-view data. Non-linear CCA extends this notion to a broader family of transformations, which are more powerful in many real-world applications. Given the joint probability, the Alternating Conditional Expectation (ACE) algorithm provides an optimal solution to the non-linear CCA problem. However, it suffers from limited performance and an increasing computational burden when only a finite number of samples is available. In this work we introduce an information-theoretic compressed representation framework for the non-linear CCA problem (CRCCA), which extends the classical ACE approach. Our suggested framework seeks compact representations of the data that allow a maximal level of correlation. This way we control the trade-off between the flexibility and the complexity of the model. CRCCA provides theoretical bounds and optimality conditions, as we establish fundamental connections to rate-distortion theory, the information bottleneck and remote source coding. In addition, it allows a soft dimensionality reduction, as the compression level is determined by the mutual information between the original noisy data and the extracted signals. Finally, we introduce a simple implementation of the CRCCA framework, based on lattice quantization.}

\keyword{ Canonical Correlation Analysis, Alternating Conditional Expectation, Remote Source Coding, Dimensionality Reduction, Information Bottleneck}

\begin{document}



\section{Introduction}
Canonical correlation analysis (CCA) seeks linear projections of  two given random vectors so that the extracted (possibly lower dimensional) variables are maximally correlated \cite{hotelling1936relations}. CCA is a powerful tool in the analysis of \textit{paired data} $(X,Y)$, where $X$ and $Y$ are two different representations of the same set of objects. It is commonly used in a variety of applications, such as speech recognition \cite{arora2012kernel}, natural language
processing \cite{dhillon2011multi}, cross-modal retrieval \cite{gong2014improving},  multimodal signal processing \cite{slaney2001facesync}, computer vision \cite{kim2007tensor} and many others. 

The CCA framework has gained a considerable amount of attention in recent years due to several important contributions (see Section \ref{previous_work}). However, one of its major drawbacks is its restriction to linear projections, whereas many real-world setups exhibit highly non-linear relationships. To overcome this limitation, several non-linear CCA extensions have been proposed. Van Der Burg and De Leeuw  \cite{van1983non} studied a non-linear CCA problem under a specific family of transformations. Breiman and Friedman \cite{breiman1985estimating} considered a generalized non-linear CCA setup, in which the transformations are not restricted to any model. They derived the optimal solution for this problem, under a known joint probability. For a finite sample size,  Akaho \cite{akaho2001kernel} suggested a kernel version of CCA (KCCA) in which non-linear mappings are chosen from two reproducing kernel Hilbert spaces (RKHS). Wang \cite{wang2007variational} and Kalmi et al. \cite{klami2013bayesian} considered a Bayesian approach to non-linear CCA and provided inference algorithms and variational approximations that learn the structure
of the underlying model. Later,  Andrew et al. \cite{andrew2013deep} introduced \textit{Deep CCA} (DCCA), where  the projections are obtained from two deep neural networks that are trained to output maximally
correlated signals. In recent years, non-linear CCA gained a renewed growth of interest in the machine learning community (for example, \cite{michaeli2016nonparametric}).  

Non-linear CCA methods are advantageous over linear CCA in a range of applications (for example, \cite{hardoon2004canonical, wang2015deep}). However, two major drawbacks typically characterize most methods. First, although there exist several studies on the statistical properties of the linear CCA problem (\cite{arora2016stochastic,arora2017stochastic}), the non-linear case remains quite unexplored with only few recent studies (for example, \cite{wang2015stochastic}).  Second, current non-parametric (and henceforth non-linear) CCA methods are typically computationally demanding. While there exist several parametric non-linear CCA methods that address this problem, completely non-parametric methods (in which the solution is not restricted to a parametric family, nor utilize parametric density estimation, as in \cite{michaeli2016nonparametric}) are often impractical to apply to large data sets.

In this work we consider a compressed representation  formulation to the non-linear CCA framework (namely, CRCCA), which demonstrates many desirable properties. Our suggested formulation regularizes the non-linear CCA problem in an explicit and theoretically sound manner. In addition, our suggested scheme drops the traditional hard dimensionality reduction of the CCA framework and replaces it with constraints on the mutual information between the given noisy data and the extracted signals.  This results in a soft dimensionality reduction, as the compression level is controlled by the constraints of our approach. 

The CRCCA framework provides theoretical bounds and optimality conditions, as we establish fundamental connections to the theory of rate-distortion (see, e.g., Chapter 10 of \cite{cover2012elements}) and the information bottleneck \cite{tishby1999information}. Given the joint probability, we achieve a coupled variant of the classical Distortion-Rate problem, where we maximize the correlation between the representation with constraints on the representation rates. Furthermore, we provide an empirical solution for the CRCCA problem, when only a finite set of samples is available. Our suggested solution is both theoretically sound and computationally efficient. A Matlab implementation of our suggested approach is publicly available at the first author's webpage\footnote{www.math.tau.ac.il/$\sim$amichaip}.

It is important to mention that the CRCCA framework can also be interpreted from a classical information theory view point. Consider two  disjoint terminals $X$ and $Y$, where each vector is to be transmitted (independently of the other) through a different rate-limited noiseless channel. Then, the CRCCA framework seeks minimal transmission rates, given a prescribed correlation between the received signals. 

The rest of this manuscript is organized as follows. In Section \ref{previous_work} we briefly review relevant concepts and previous studies of the CCA problem. In Section \ref{problem_formulation} we formally introduce our suggested CRCCA framework. Then, we describe our suggested solution in Section \ref{Iterative_projections_solution}. In Section \ref{From_theory_to_practice} we drop the known probability assumption and discuss the finite sample-size regime. We conclude with a series of synthetic and real-world experiments in Section \ref{experiments}.

\section{Previous Work}
\label{previous_work}
Let $X\in \mathbb{R}^{d_X}$ and $Y\in \mathbb{R}^{d_Y}$ be two random vectors. For the simplicity of the presentation we assume that $X$ and $Y$ are also zero-mean. The CCA framework seeks two transformations, $U=\phi(X)$ and $V=\psi(Y)$ such that
\begin{equation} \label{CCA}
\begin{aligned}
& \max_{\substack{U=\phi(X)\\ V=\psi(Y)}}
& & \sum_{i=1}^d \mathbb{E}(U_i V_i) \\
& \text{subject to}
& & \mathbb{E}(U)=\mathbb{E}(V)=0\\
&
& & \mathbb{E}(UU^T)=\mathbb{E}(VV^T)=I\\
\end{aligned}
\end{equation}
where $d\leq \min{(d_X,d_Y)}$. Notice that the notation $U=\phi(X)$ implies that all $X$ variables are transformed simultaneously by a single multivariate transformation $\phi$. We refer to (\ref{CCA}) as CCA, where linear CCA \cite{hotelling1936relations}, is under the assumption that  $\phi(\cdot)$ and $\psi(\cdot)$ are linear ($U=A X$ and $V=B Y$). 

The solution to the linear CCA problem can be obtained from the singular value decomposition of the matrix $\Sigma_X^{-\frac{1}{2}}\Sigma_{XY}\Sigma_Y^{-\frac{1}{2}}$, where $\Sigma_X, \Sigma_Y$ and $\Sigma_{XY}$ are the covariance matrices of $X$, $Y$ and the cross-covariance of $X$ and $Y$, respectively. 
In practice, the covariance matrices are typically replaced by their empirical estimates,  obtained from a finite set of samples. The linear CCA has been studied quite extensively in recent years, and has gained popularity due to several contributions.  Ter Braak \cite{ter1990interpreting} and Graffelman \cite{graffelman2005enriched} showed that CCA can be enhanced with powerful graphics called biplots. 
Witten et al. \cite{witten2009penalized} introduced a regularized variant of CCA, which improves generalization. This method was further extended to a sparse CCA setup  \cite{witten2009penalized, witten2009extensions}. More recently, Graffelman et al. \cite{graffelman2018exploration} adapted a compositional setting for linear CCA using non-linear log-ratio transformations.

Non-linear CCA is a natural extension to the linear CCA problem. Here $\phi$ and $\psi$ are not restricted to be linear projections of $X$ and $Y$. This problem was first introduced by  Lancaster \cite{lancaster1958structure} and Hannan \cite{hannan1961general} and was later studied in different setups, under a variety of models (for example, \cite{van1983non, van1994overals}). It is important to emphasize that non-linear CCA may also be viewed as a nonlinear multivariate analysis technique, with several important theoretical and algorithmic contributions \cite{gifi1990nonlinear}. A major milestone in the study non-linear CCA  was achieved by Breiman and Friedman \cite{breiman1985estimating}. In their work, Breiman and Friedman showed that the optimal solution to (\ref{CCA}), for $d=1$,  may be obtained by a simple alternating conditional expectation procedure, denoted ACE. Their results were later extended to any $d\leq \min{(d_X,d_Y)}$, as shown, for example, in \cite{makur2015efficient}. Here, we briefly review the ACE framework.

Let us begin with the first set of components, $i=1$. Assume that $V_1=\psi_1(Y)$ is fixed, known and satisfies the constraints. Then, the optimization problem (\ref{CCA}) is only with respect to $\phi_1$ and by Cauchy-Schwarz inequality, we have that
\begin{align}
\mathbb{E}(U_1 V_1)=&\mathbb{E}_X \left(\phi_1(X)\mathbb{E}(\psi_1(Y)|X)\right) \leq \sqrt{\text{var}(\phi_1(X))} \sqrt{\text{var}(\mathbb{E}(\psi_1(Y)|X))} 
\end{align}
with equality if and only if $\phi_1(X)=c\cdot \mathbb{E}(\psi_1(Y)|X)$.  Therefore, choosing a constant $c$ to satisfy the unit variance constraint we achieve $\phi_1(X)={\mathbb{E}(\psi_1(Y)|X)}/{\sqrt{var(\mathbb{E}(\psi_1(Y)|X))}}$. In the same manner we may fix $\phi_1(X)$ and attain $\psi_1(Y)={\mathbb{E}(\phi_1(X)|Y)}/{\sqrt{var(\mathbb{E}(\phi_1(X)|Y))}}$. These coupled equations are in fact necessary conditions for the optimality of $\phi_1$ and $\psi_1$, leading to an alternating procedure in which at each step we fix one transformation and optimize with respect to the other. Once $\phi_1$ and $\psi_1$ are derived, we continue to the second set of components, $i=2$, under the constraints that they are uncorrelated with $U_1,V_1$. This procedure continues for the remaining components.  Breiman and Friedman \cite{breiman1985estimating} proved that ACE converges to the global optimum.  In practice, the conditional expectations are estimated from training data  $\{x_i, y_i\}_{i=1}^{n}$ using nonparametric regression, usually in the form of $k$-nearest neighbors ($k$-NN). Since this computationally demanding step has to be executed repeatedly, ACE and its extensions are impractical for large data analysis.

\textit{Kernel CCA} (KCCA) is an alternative non-linear CCA framework \cite{lai2000kernel,akaho2001kernel, uurtio2019large, yu2019accelerated}. In KCCA , $\phi \in \mathcal{A}$ and $\psi \in \mathcal{B}$, where  $\mathcal{A}$ and $\mathcal{B}$ are two reproducing kernel Hilbert spaces (RKHSs) associated with user-specified kernels $k_x(\cdot, \cdot)$ and $k_y(\cdot, \cdot)$. By the representer theorem \cite{scholkopf2001generalized}, the projections can be written in terms of the training samples, $\{x_i, y_i\}_{i=1}^{n}$, as $U_i=\sum_{j=1}^n a_{ji}k_x(X,x_j)$ and $V_i=\sum_{j=1}^n b_{ji}k_y(Y,y_j)$ for some coefficients $a_{ji}$ and $b_{ji}$. Denote the kernel matrices as $K_x=[k_x(x_i,x_j)]$ and $K_y=[k_y(y_i,y_j)]$. Then, the optimal
coefficients are computed from the eigenvectors of the matrix $(K_x+r_xI)^{-1}Ky(K_y+r_yI)^{-1}Kx$ where $r_x$ and $r_y$ are positive parameters. Computation of the exact solution is intractable for large datasets due to the memory cost of storing the kernel matrices and the time complexity of solving dense eigenvalue systems. To address this caveat, Bach and Jordan \cite{bach2002kernel} and Hardoon et al. \cite{hardoon2004canonical} suggested several low-rank matrix approximations. Later, Halko et al. \cite{halko2011finding} considered randomized singular value decomposition (SVD) methods to further reduce the computational burden of  KCCA. Additional modifications were  explored by Arora and Livescu \cite{arora2012kernel}. 

More recently, Andrew et al. \cite{andrew2013deep} introduced \textit{Deep CCA} (DCCA). Here, $\phi \in \mathcal{A}$ and $\psi \in \mathcal{B}$, where  $\mathcal{A}$ and $\mathcal{B}$ are families of functions that can be implemented using two \textit{Deep Neural Networks} (DNNs) of predefined architectures. As many DNN frameworks, DCCA is a scalable solution  which demonstrates favorable generalization abilities in large data problems. Wang et al. \cite{wang2016deep} extended DCCA by introducing autoencoder regularization terms, implemented by additional DNNs. Specifically, Wang et al. \cite{wang2016deep} maximize the correlation between $U=\phi(X)$ and $V=\psi(Y)$ where $\phi,\psi$ are DNNs (similarly to DCCA), while regulating the squared reconstruction error, $||X-\tilde{\phi}(U)||^2_2$ and $||Y-\tilde{\psi}(V)||^2_2$ where $\tilde{\phi}$ and $\tilde{\psi}$ are additional DNNs, optimized over possibly different architectures than $\phi$ and $\psi$. Wang et al. \cite{wang2015deep} called this method \textit{Deep Canonically Correlated Autoencoders} (DCCAE) and demonstrated its abilities in a variety of large scale problems. Importantly, they showed that the additional regularization terms improve upon the original DCCA framework as they regulate its flexibility. Unfortunately, as most artificial neural network methods, both DCCA and DCCAE provide a limited understanding of the problem. 

\section{Problem Formulation}
\label{problem_formulation}
Let $X\in \mathbb{R}^{d_X}$ and $Y\in \mathbb{R}^{d_Y}$ be two random vectors. For the simplicity of the presentation we assume that $d_X=d_Y=d$. It is later shown that our derivation holds for any $d_X \neq d_Y$ and $d\geq0$. 
 Let $\phi: \mathbb{R}^{d_X} \rightarrow \mathbb{R}^{d}$ and $\psi: \mathbb{R}^{d_y} \rightarrow \mathbb{R}^{d}$ be  two transformations.  Let $U=\phi(X)$ and $V=\psi(Y)$ be two vectors in $\mathbb{R}^d$. Notice that $\phi$ and $\psi$ are not necessarily deterministic transformations, in the sense that the conditional distributions $p(u|x)$ and $p(v|y)$ may be non-degenerate distributions. In this work we generalize the classical CCA formulation (\ref{CCA}), as we impose additional mutual information constraints on the transformations that we apply. Specifically, we are interested in $U$ and $V$ such that 
\begin{equation} \label{soft_CCA}
\begin{aligned}
& \max_{\substack{U=\phi(X)\\ V=\psi(Y)}}
& & \sum_{i=1}^d \mathbb{E}(U_i V_i) \\
& \text{subject to}
& & \mathbb{E}(U)=\mathbb{E}(V)=0\\
&
& & \mathbb{E}(UU^T)=\mathbb{E}(VV^T)=I\\
&
& & I(X;U) \leq R_U, \quad I(Y;V) \leq R_V\\
\end{aligned}
\end{equation}
for some fixed $R_U$ and $R_V$, where $I(X;Y)=\int_{x,y} p(x,y) \log \frac{p(x,y)}{p(x)p(y)}dxdy$ is the mutual information of $X$ and $Y$ and $p(x,y)$ is the joint distribution of $X$ and $Y$. The mutual information constraints regulate the transformations that we apply, so that in addition to maximizing the sum of correlations (as in  (\ref{CCA})), $U$ and $V$ are also restricted to be compressed representations of $X$ and $Y$, respectively. In other words, $R_U$ and $R_V$ define the amount of information preserved from the original vectors.  Notice that as $R_U$ and $R_V$ grow, (\ref{soft_CCA}) degenerates back to (\ref{CCA}). Further, it is important to emphasize that while $\phi$ and $\psi$ are not necessarily deterministic, most applications do impose such a restriction. In this case, the mutual information constraints are unbounded if $U$ and $V$ take values over a finite support. In this case, $I(X;U)=H(U)$ and $I(Y;V)=H(V)$, where $H(U)=-\sum_u p(u)\log p(u)$ is the entropy of $U$.   

The use of mutual information as a regularization term is one of the corner stones of information theory. The \textit{Minimum Description Length} (MDL) principle \cite{rissanen1978modeling} suggests that the best representation of a given set of data is the one that leads to a minimal coding length. This idea has inspired the use of mutual information as a regularization term in many learning problems, mostly in the context of rate distortion (Chapter 10 of \cite{cover2012elements}), the information bottleneck framework \cite{tishby1999information}, and different representation learning problems \cite{chigirev2003optimal}. Recently, Vera et al. \cite{vera2018role} showed that a mutual information constraint explicitly controls the generalization gap when considering a cross-entropy loss function. We further discuss the desirable properties of the mutual information constraints in Section \ref{regularization}.

Notice that the regularization terms in the DCCAE framework (discussed above) is related to our suggested mutual information constraints. However, it is important to emphasize the difference between the two. The autoencoders in the DCCAE framework suggest an explicit reconstruction architecture, which strives to maintain a small reconstruction error with the original representation. On the other hand, our mutual information constraints regulate the ability to reconstruct the original representation. In other words, our suggested framework restricts the amount of information that is preserved with the original representation while DCCAE minimizes the reconstruction error with the original representation. 






We refer to our constrained optimization problem (\ref{soft_CCA}) as \textit{Compressed Representation CCA} (CRCCA). Notice that traditionally, CCA refers to linear transformations. Here we again consider CCA in the wider sense, as the transformations may be non-linear and even non-deterministic. Further, notice that we may interpret  (\ref{soft_CCA}) as a soft version of the CCA problem. In the classical CCA setup, the applied transformations strive to maximize the correlations and rank them in a descending order.  This implicitly suggests a hard  dimensionality reduction, as one may choose subsets of components of $U$ and $V$ that have the strongest correlations. In our formulation (\ref{soft_CCA}), the mutual information constraints allow a soft dimensionality reduction; while $X$ and $Y$ are transformed to maximize the objective, the transformations also compress $X$ and $Y$ in the classical  rate-distortion sense. For example, assume that the transformations are deterministic. Then $I(X;U)=H(U)$ and $I(Y;V)=H(V)$ (as discussed above).  Here, (\ref{soft_CCA}) may be interpreted as a correlation maximization problem, subject to a constraint on the maximal number of bits allowed to represent (or store) the resulting representations. In the same sense, the classical CCA formulation imposes a hard dimensionality reduction, as it constraints the number of dimensions allowed to represent (or store) the new variables.       In other words, instead of restricting the number of dimensions allowed to represent the variables, we restrict the level of information allowed to represent them. We emphasize this idea in Section \ref{experiments}.

\section{Iterative Projections Solution}
\label{Iterative_projections_solution}
Inspired by Breiman and Friedman \cite{breiman1985estimating} we suggest an iterative approach for the CRCCA problem (\ref{soft_CCA}). Specifically, in each iteration, we fix one of the transformations and maximize the objective with respect to the other. Let us illustrate our suggested approach as we fix $V$ and maximize the objective with respect to $U$. 

First, notice that our objective may be compactly written  as $\sum_{i=1}^d \mathbb{E}(U_i V_i)=\mathbb{E}(V^T U)$. Since $\mathbb{E}(VV^T)$ is fixed (and our constraint suggests that $\mathbb{E}(UU^T)=I$), we have that maximizing $\sum_{i=1}^d \mathbb{E}(U_i V_i)$ is equivalent to minimizing $\mathbb{E}||U-V||^2$. Therefore, the basic step in our iterative procedure is 
\begin{equation} \label{on_side_CCA}
\begin{aligned}
& \min_{p(u|x)}
& & \mathbb{E}||U-V||^2 \\
&  \; \text{s.t.}
& &  I(X;U) \leq R_U, \; \mathbb{E}(U)=0, \; \mathbb{E}(UU^T)=I,
\end{aligned}
\end{equation}
or equivalently (as in rate-distortion theory \cite{cover2012elements})
\begin{equation} \label{noisy_source_coding}
\begin{aligned}
& \min_{p(u|x)}
& & I(X;U) \\
& \; \text{s.t.}
& &  \mathbb{E}||U-V||^2 \leq D, \;\mathbb{E}(U)=0, \; \mathbb{E}(UU^T)=I.
\end{aligned}
\end{equation}
This problem is widely known in the information theory community as remote/noisy source coding (\cite{dobrushin1962information,wolf1970transmission}) with  additional constraints on the second order statistics of $U$. 
Therefore, {our suggested method provides a local optimum to (\ref{soft_CCA}) by iteratively solving a remote source coding problem, with additional second order statistics constraints}.  

Remote source coding is a variant of the classical source coding (rate-distortion) problem  \cite{cover2012elements}. Let $V$ be a remote source that is unavailable to the encoder. Let $X$ be a random variable that depends of $V$ through a (known) mapping $p(x|v)$, and is available to the encoder. The remote source coding problem seeks the minimal possible compression rate of $X$, given a prescribed maximal reconstruction error of $V$ from the compressed representation of $X$. Notice that for $V=X$, the remote source coding problem degenerates back to the classical source coding regime. Remote source coding has been extensively studied over the years. Dobrushin and Tsybakov \cite{dobrushin1962information}, and later Wolf and Ziv \cite{wolf1970transmission}, showed that the solution to this remote source coding problem (that is, the optimization problem in (\ref{noisy_source_coding}), without the second order statistics constraints) is achieved by a two step decomposition. First, let  $\tilde{V}=\mathbb{E}(V|X)$ be the conditional expectation of $V$ given $X$, which defines the optimal minimum mean square error (MMSE) estimator of the remote source $V$ given the observed $X$. Then, $U$ is simply the rate-distortion solution with respect to $\tilde{V}$. It is immediate to show that the same decomposition holds for our problem, with the additional second order statistic constraints. In other words, in order to solve (\ref{noisy_source_coding}), we first compute $\tilde{V}=\mathbb{E}(V|X)$, followed by 
\begin{equation}
\begin{aligned}
\label{rate_distortion}
& \min_{p(u|\tilde{v})} \; I(\tilde{V};U) \\  
& \; \text{s.t.} \;\; \;\;\mathbb{E}||U-\tilde{V}||^2 \leq D, \; \mathbb{E}(U)=0, \; \mathbb{E}(UU^T)=I. 
\end{aligned}
\end{equation}
We notice that (\ref{rate_distortion}) is simply the rate distortion function $\tilde{V}$ for square error distortion, but  with the additional (and untraditional) constraints on the second order statistics of the representation.  

 \subsection{Optimality Conditions}
\label{optimality_conditions}
Let us now derive the optimality conditions for each step of our suggested iterative projection algorithm. For this purpose, we assume that the joint probability distribution of $X$ and $Y$ is known. For the simplicity of the presentation we focus on the one dimensional case, $X,Y,U,V \in R$. As we do not restrict ourselves to deterministic transformations, the solution to (\ref{rate_distortion}) is fully characterized by the conditional probability $p(u|\tilde{v})$. 

\begin{Lemma}
\label{lemma_optimality_conditions}
In each step of our suggested iterative projections method, the optimal transformation (\ref{on_side_CCA}) must satisfy the optimality conditions of (\ref{rate_distortion}): 
\begin{enumerate}
  \item $p(u|\tilde{v})=p(u)e^{-\tilde{\lambda}(\tilde{v})}e^{-\eta(u-\tilde{v})^2-\tau u-\mu u^2}$
  \vspace{0.1cm}
  \item $p(u)=\int_{\tilde{v}}p(u|\tilde{v})p(\tilde{v})d\tilde{v}$
\end{enumerate}
 where $\tilde{V}=\mathbb{E}(V|X)$,  $\tilde{\lambda}(\tilde{v})=1-\frac{\lambda(\tilde{v})}{p(\tilde{v})}$
and $\eta, \tau, \mu, \lambda(\tilde{v})$ are the Lagrange multipliers associated with the constraints of the problem.
\end{Lemma}

A proof of this Lemma is provided in Appendix A. As expected, these conditions are identical to the Arimoto-Blahut equations (Chapter 10 of \cite{cover2012elements}), with the additional term $e^{-\tau u-\mu u^2}$ that corresponds to the second order statistics constraints. This allows us to derive an iterative algorithm (Algorithm \ref{alg}), similar to Arimoto-Blahut, in order to find the (locally) optimal mapping $p(u|\tilde{v})$, with the same (local) convergence guarantees as Arimoto-Blahut. Our suggested algorithm is also highly related to the iterative approach of the information bottleneck solution, in the special Gaussian case \cite{chechik2005information}. 

\begin{algorithm} 
\caption{Arimoto Blahut pseudo-code for rate distortion with second order statistics constraints}
\begin{algorithmic} [1]
\setstretch{1.7}
\REQUIRE $p(\tilde{v})$
\ENSURE Fix $p(u), \eta, \tau, \mu, \lambda(\tilde{v})$
\STATE  Set $\tilde{\lambda}(\tilde{v})=1-{\lambda(\tilde{v})}/{p(\tilde{v})}$
\STATE  Set $p(u|\tilde{v})=p(u)e^{-\tilde{\lambda}(\tilde{v})}e^{-\eta(u-\tilde{v})^2-\tau u-\mu u^2}$ \label{first_step}
\STATE Set $p(u)=\int_{\tilde{v}}p(u|\tilde{v})p(\tilde{v})d\tilde{v}$
\STATE  Set  $\tau$ so that $\mathbb{E}(U)=0$
\STATE   Set  $\mu$ so that $\mathbb{E}(U^2)=1$
\STATE  Go to Step \ref{first_step} until convergence
\end{algorithmic}
\label{alg}
\end{algorithm} 

Generalizing the optimality conditions (and the corresponding iterative algorithm) to the vector case is straight forward (see Appendix A). Again, the Lagrangian leads to Arimoto-Blahut equations, with an additional exponential term, $e^{-\tau^T u-u^T\mu u}$. Here, $\tau$ is a vector of Lagrangian multipliers while $\mu$ is a matrix.

\section{Compressed Representation CCA for Empirical Data}
\label{From_theory_to_practice}
To this point, we considered the CRCCA problem under the assumption that the joint probability distribution of $X$ and $Y$ is known. However, this assumption is typically invalid in real world setups. Instead, we are given a set of i.i.d. samples $\{x_i,y_i\}_{i=1}^n$ from $p(x,y)$. We show that in this setup, Dobrushin and Tsybakov optimal decomposition \cite{dobrushin1962information} may be redundant, as we attempt to directly solve (\ref{noisy_source_coding}). 

\subsection{Previous Results} 

Let us first revisit the iterative projections solution (Section \ref{Iterative_projections_solution}) in a real-world setting. Here, in each iteration we are to consider an empirical version of  (\ref{noisy_source_coding}). This problem is equivalent to the empirical remote source coding problem (up to the additional second order statistics constraints), which was first studied by Linder et al. \cite{linder1997empirical}. In their work, Linder et al. followed Dobrushin and Tsybakov \cite{dobrushin1962information}, and decomposed the problem to conditional expectation estimation (denoted as $\hat{\mathbb{E}}(V|X)$) followed by empirical vector quantization, ${Q}(\cdot)$. They showed that under an additive noise assumption ($X=V+\epsilon$), the convergence rate of the empirical distortion is
\begin{align}
\label{linder_bound}
\mathbb{E}||{Q}^*(\hat{\mathbb{E}}(V|X))-V||^2\leq& D^*_N+8B\sqrt{\frac{2 d_x N \log n}{n}}+O\left(n^{-\frac{1}{2}}\right)+8\sqrt{Be_n}+e_n
\end{align}
where ${Q}^*$ is the optimal empirically trained $N$-level vector quantizer, $D^*_N$ is the distortion of the optimal $N$-level vector quantizer for the remote source problem (where the joint distribution is known), $B$ is a known constant that satisfies $P(||X||^2\leq B)=1$, 
and $e_n$ is the mean square error of the empirical conditional expectation, $\mathbb{E}||\mathbb{E}(V|X)-\hat{\mathbb{E}}(V|X)||^2=e_n$. Notice that this  bound has three major terms:  the irreducible quantization error $D^*_N$, the conditional expectation estimation error $e_n$, and the empirical vector quantization term, $8B\sqrt{\frac{2 d_x N \log n}{n}}$. Although their analysis focuses on vector quantization, it is evident that (\ref{linder_bound}) strongly depends on the performance of the conditional expectation estimator $\hat{\mathbb{E}}(V|X)$. In fact, if we choose a non-parametric  $k$-nearest neighbors estimator (like \cite{breiman1985estimating}), we have that $\mathbb{E}||\mathbb{E}(V|X)-\hat{\mathbb{E}}(V|X)||^2=O\left(n^{-\frac{2}{d+2}}\right)$ \cite{gyorfi2006distribution}, which is significantly worse than the rate imposed by the empirical vector quantizer.  An additional drawback of (\ref{linder_bound}) is the restrictive additive noise modeling assumption. For more general cases, Gy{\"o}rfi and Wegkamp \cite{gyorfi2008quantization} derived similar convergence rates under broader sub-Gaussian models.  

In our work we take a different approach, as we drop Dobrushin and Tsybakov decomposition and attempt to solve (\ref{noisy_source_coding}) directly. Importantly, since both the $k$-nearest neighbors and the vector quantization modules require a significant computational effort as the dimension of the problem increases, we take a more practical approach and design (a variant of) a lattice quantizer. 

\subsection{Our Suggested Method}
\label{our_suggested_method}
As previously discussed, Dobrushin and Tsybakov decomposition yields an unnecessary statistical and computational burden, as we first estimate the conditional expectation, and then use it as a plug-in for the statistic that we are essentially interested in.  Here, we drop this decomposition and solve (\ref{noisy_source_coding}) directly. For this purpose we apply a remote source variant of lattice quantization. 

\subsubsection{Lattice Quantization}
Lattice quantization \cite{zamir2014lattice} is a popular alternative for optimal vector quantization. In this approach, the partitioning of the quantization space is known and predefined. Given a set of observations $\{x_i\}_{i=1}^n$, the fit (also called \textit{representer} or \textit{centroid}) of each quantization cell is simply the average of all the $x_i$'s that are sampled in that cell (see the left chart of Figure \ref{quantizers}).  We denote the (empirically trained) fixed lattice quantizer of $X$ as ${U}_{LQ}(X)$. Computationally, Lattice quantizers are scalable to higher dimensions, as they only require a fixed partitioning of the support of $X\in \mathbb{R}^{d_x}$. Further, it can be shown that the rate of the quantizer (which corresponds to $I(X,{U}_{LQ}(X))$) is simply the entropy of ${U}_{LQ}(X)$ \cite{ziv1985universal}. This allows a simple way to verify the performance of the lattice quantizer. In addition, it quantifies the number of bits required to encode ${U}_{LQ}(X)$. Lattice quantizers followed by entropy coding asymptotically ($n\rightarrow \infty$) approach  the rate distortion bound, in high rates (Chapter 7 of \cite{zamir2009lattices}).  Interestingly, it can be shown  that in low rates, the optimal (dithered) lattice quantizer is up to half a bit worse than the rate distortion bound, while the uniform dithered quantizer is up to $0.754$ bits worse than the rate-distortion bound \cite{zamir1992universal}. The performance of a lattice quantizer may be further improved with pre/post filtering \cite{zamir1996information}.  

\subsubsection{CRCCA by Quantization}
In our empirical version of (\ref{noisy_source_coding}) we are given a set of observations $\{x_i,v_i  \}_{i=1}^n$. Define the set of all possible quantizer of $X$ as $\mathcal{Q}(X)$. We would like to find a quantizer $U_q \in \mathcal{Q}(X)$, such that
\begin{equation} \label{empirical_on_side_CCA}
\begin{aligned}
& \min_{U_q \in \mathcal{Q}(X)}
& & \mathbb{E}||U_q-V||^2 \\
&  \; \text{s.t.}
& &  H(U_q) \leq R_U, \; \mathbb{E}(U_q)=0, \; \mathbb{E}(U_q U_q^T)=I.
\end{aligned}
\end{equation}
As a first step towards this goal, let us define a simpler problem. Denote the set of all uniform and fixed lattice quantizer of $X$ by $\mathcal{Q}_U(X)$. Then, the \textit{remote source uniform quantization} problem is defined as 
\begin{equation} \label{RSUQ}
\begin{aligned}
& \min_{U_q \in \mathcal{Q}_U(X)}
& & \mathbb{E}||U_q-V||^2 \\
&  \; \text{s.t.}
& &  H(U_q) \leq R_U.
\end{aligned}
\end{equation}

\begin{figure}[ht]
\centering
\includegraphics[width =0.6\textwidth,bb= 140 220 660 420,clip]{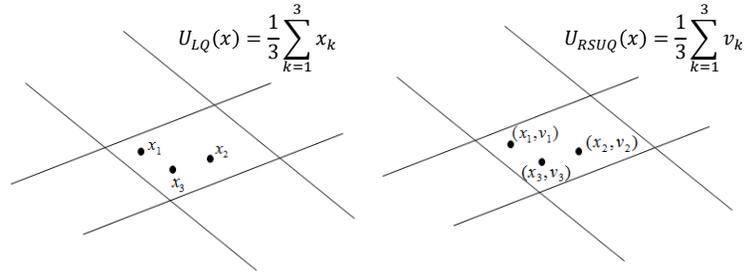}
\caption{Examples of uniform quantization. Left: classical uniform quantization. Right: remote source uniform quantization.}
\label{quantizers}
\vspace{-1.5em}
\end{figure}

In other words, (\ref{RSUQ}) is minimized over a subset of quantizers  $\mathcal{Q}_U(X) \subset \mathcal{Q}(X)$, and drops the second order statistics constraints that appear in (\ref{empirical_on_side_CCA}). In order to solve (\ref{RSUQ}), we follow the lattice quantization approach described above; we first apply a (uniform and fixed) partitioning on the space of $X$. Then, the fit of each quantization cell is simply the average of all $v_i$'s that correspond to the $x_i$'s that were sampled in that cell (see the right chart of Figure \ref{quantizers} for example). We denote this quantizer as ${U}_{RSUQ}(X)$, where \textit{RSUQ} stands for \textit{remote source uniform quantization}. Notice that our suggested partitioning is not an optimal solution to  (\ref{empirical_on_side_CCA}) (even without the second order statistics constraints), as we apply a simplistic uniform quantization to each dimension. However, it is easy to verify that  ${U}_{RSUQ}(X)$ is the empirical minimizer of (\ref{RSUQ}). Finally, in order to satisfy the second order constraints, we apply a simple linear transformation, ${U}_q=A{U}_{RSUQ}(X)+B$. Lemma \ref{second_order_statistics_lemma} below shows that  ${U}_q=A{U}_{RSUQ}(X)+B$ is indeed the empirical risk minimizer of  (\ref{RSUQ}), with the additional second order statistics constraints. 

\begin{Lemma}
\label{second_order_statistics_lemma}
Let ${U}_{RSUQ}(X)$ be the remote source uniform quantizer that is the empirical risk minimizer of (\ref{RSUQ}). Then, $A{U}_{RSUQ}(X)+B$ is the remote source uniform quantizer that  is the empirical risk minimizer of (\ref{RSUQ}), with the additional constraints, for some constant matrix $A$ and a vector $B$.
\end{Lemma}
A proof for lemma \ref{second_order_statistics_lemma} is provided in Appendix B. To conclude, our suggested quantizer ${U}_q$ is the empirical minimizer of  (\ref{RSUQ}), which approximates (\ref{empirical_on_side_CCA}) over a subset of quantizers $\mathcal{Q}_U(X) \subset \mathcal{Q}(X)$. Algorithm \ref{alg2} summarizes our suggested approach.

\begin{algorithm} 
\caption{A single step of CRCCA by quantization}
\begin{algorithmic} [1]
\setstretch{1.7}
\REQUIRE $\{x_i,v_i\}_{i=1}^n$, a fixed uniform quantizer $Q(x)$
\STATE  Set $\mathcal{I}(\alpha)= \left\{j\;| \; Q(x_j)=\alpha\right\}$, the indices of the samples  that are mapped to the quantization cell $\alpha$
\STATE Set ${U}_{RSUQ}(x_i)=\frac{1}{|\mathcal{I}(Q(x_i))|}\sum_{k \in \mathcal{I}(Q(x_i))}v_k$
\STATE  Set  ${U}_q(x_i)=A{U}_{RSUQ}(x_i)+B$ such that $\sum_{i=1}^n {U}_q(x_i)=0$ and  $\sum_{i=1}^n {U}_q(x_i) {U}^T_q(x_i)=I$
\end{algorithmic}
\label{alg2}
\end{algorithm}

Notice that the uniform quantization scheme described above may also be viewed as a \textit{partitioning estimate} of the conditional expectation \cite{gyorfi2006distribution}, where the number of partitions is predetermined by the prescribed quantization level. The partitioning estimate is a local averaging estimate such that for a given $x$ it takes the average of those $v_i$'s for which $x_i$ belongs to the same cell as $x$. Formally,   
$$\hat{\mathbb{E}}(V|X=x)=\frac{\sum_{i=1}^{n}v_i\mathbbm{1}\left\{Q(x_i)=Q(x)\right\}}{\sum_{i=1}^{n}\mathbbm{1}\left\{Q(x_i)=Q(x)\right\}}.$$

\subsubsection{Convergence Analysis}  
The partitioning estimate holds several desirable properties. Gy{\"o}rfi et al.  \cite{gyorfi2006distribution} showed that it is (weakly) universally consistent. Further, they showed that under the assumptions that $X$ has a compact support $\mathcal{S}$, the conditional variance is bounded, $\text{var} (V|X=x) \leq \sigma^2$,  and the conditional expectation is smooth enough, $|\mathbb{E}(V|X=x_1)-\mathbb{E}(V|X=x_2)| \leq C||x_1-x_2||$, then the partitioning estimate rate of convergence follows
\begin{align}
&\mathbb{E}||\hat{\mathbb{E}}(V|X)-\mathbb{E}(V|X)||^2 \leq \hat{c}\frac{\sigma^2+\text{sup}_{x \in \mathcal{S}}|\mathbb{E}(V|X=x)|^2}{nh_n^d}+d\cdot C^2 \cdot h^2_n 
\end{align}
where $\hat{c}$ depends only on $d$ and the diameter of $S$, and $h_n^d$ is the volume of the cubic cells. Thus, for 
\begin{equation}\label{cubic_cell}
h_n=c' \left(   \frac{\sigma^2+\text{sup}_{x \in S}|\mathbb{E}(V|X=x)|^2}{C^2}  \right)^{1/(d+2)}n^{-\frac{1}{d+2}}
\end{equation} we have that 
\begin{equation}
\label{roc}
 \mathbb{E}||\hat{\mathbb{E}}(\mathbb{E}|X)-\mathbb{E}(V|X)||^2 \leq\zeta(\sigma,\mathcal{S},d,C)\cdot n ^{-\frac{2}{d+2}}
\end{equation}
where $$\zeta(\sigma,\mathcal{S},d,C)=c''\left( \sigma^2+\text{sup}_{x \in \mathcal{S}}|\mathbb{E}(V|X=x)|^2 \right)C^{\frac{2d}{d+2}}.$$ 
We omit the description of some of the constants, as they appear in detail in \cite{gyorfi2006distribution}. The derivation above suggests that for a choice of a cubic volume which follows (\ref{cubic_cell}), the rate of convergence of the partition estimate is asymptotically identical to the $k$-nearest neighbors estimate. This shall not come as a surprise, since both estimates apply a non-parametric estimation that performs a local averaging. However, the partition estimate is much simpler to apply in practice, as previously discussed. 

Finally, by applying a partitioning estimate in each step of our iterative projections solution, we show that under the same assumptions made by Gy{\"o}rfi et al.  \cite{gyorfi2006distribution}, and a choice of cubic cell volume as in (\ref{cubic_cell}), we have that 
\begin{align}
\mathbb{E}||{U}_{RSUQ}(X)-V||^2=&\mathbb{E}||V-\mathbb{E}(V|X)||^2+\mathbb{E}||{U}_{RSUQ}(X)-\mathbb{E}(V|X)||^2\leq\\\nonumber
&\mathbb{E}||V-\mathbb{E}(V|X)||^2+\zeta(\sigma,S,d,C) \cdot n^{-\frac{2}{d+2}}
\end{align}
where $\mathbb{E}||V-\mathbb{E}(V|X)||^2$ is the irreducible error and the second inequality is due to the convergence rate of the partitioning estimate, as appears in (\ref{roc}) . As we compare this result to (\ref{linder_bound}), we notice that while the rate of convergence is asymptotically equivalent (assuming a $k$-NN estimate in (\ref{linder_bound})), our result is not restricted to additive noise models (or any other noise model). In addition, Linder et al. bound  (\ref{linder_bound}) converges to $D^*_N$, the distortion of the optimal $N$-level vector quantizer. Alternatively, our convergence is to the irreducible error. 

It is important to emphasize that at the end of the day, our suggested method replaces the choice of $k$-NN estimate (as in ACE) with a partitioning estimate, where both estimates are known to be highly related. However, our suggested approach provides a sound information-theoretic justification that allows additional analytical properties, computational benefits and performance bounds. In addition, it results in an implicit regularization, as discussed below.

\subsection{Regularization}
\label{regularization}
The mutual information constraints of the CRCCA problem (\ref{soft_CCA}) hold many desirable properties. In the previous sections we focused on the information-theoretic interpretation of the problem. However, these constraints also implicitly apply regularization to the non-linear CCA problem, and by that, improve its generalization performance. Intuitively speaking, the mutual information constraint $I(X;U) \leq R_U$ suggests that $U$ is a compressed  representation of $X$. This means that by restricting the information that $U$ carries on $X$, we force the transformation to preserve only the relevant parts of $X$ with respect to the canonical correlation objective. In other words, while the objective strives to fit the best transformations to a given train-set, the mutual information constraints regularize this fit by restricting its statistical dependence with the train-set. The use of mutual information as a regularization term in learning problems is not new. The \textit{Minimum Description Length} (MDL) principle \cite{rissanen1978modeling} suggests that the best hypothesis for a given set of data is the one that leads to minimal code length needed to represent of the data. Another example is the  information bottleneck framework \cite{tishby1999information}. There, the objective is to maximize the mutual information between the labels $Y$ and a new representation of the features $T(X)$, subject to the constraint $I(X;T(X))\leq R_U$. As in our case, the objective strives to find the best fit (according to a different loss function), while the constraints serve as regularization terms. 

As demonstrated in the previous sections, the CRCCA formulation may be implemented in practice using uniform quantizers. Here, the entropy of the quantization cells replaces the mutual information constraints in regularizing the objective. A small number of cells, which typically corresponds to a lower entropy, implies that more observations are averaged together, hence more bias (and less variance). On the other hand, a greater number of cells results in a smaller number of observations that contribute to each fit, which leads to more variance (and less bias). This way, the entropy of the cells governs the well-known bias-variance trade-off.  

As before, we notice that the $k$-NN estimate may also control this trade-off, as the parameter $k$ defines the number of observations to be averaged for each fit. However, while this parameter is internal to this specific conditional expectation estimator, the number of cells in the CRCCA formulation is an explicit part of the problem formulation. In other words, CRCCA defines an explicit regularization framework to the non-linear CCA problem. 

\section{Experiments}
\label{experiments}
We now demonstrate our suggested approach in synthetic and real-world experiments. 
\subsection{Synthetic Experiments}
\begin{figure}[ht]
\centering
\includegraphics[width =0.6\textwidth,bb= 40 260 550 520,clip]{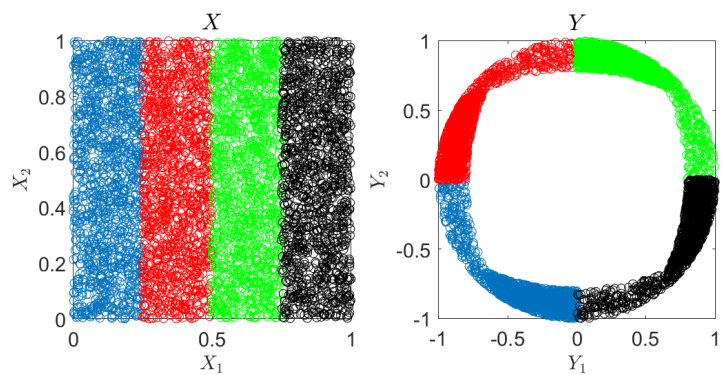}
\caption{Visualization experiment. Samples of $X$ and $Y$.}
\label{Circle_experiment}
\end{figure}

\begin{figure}[ht]
\centering
\includegraphics[width =0.6\textwidth,bb= 120 100 670 510,clip]{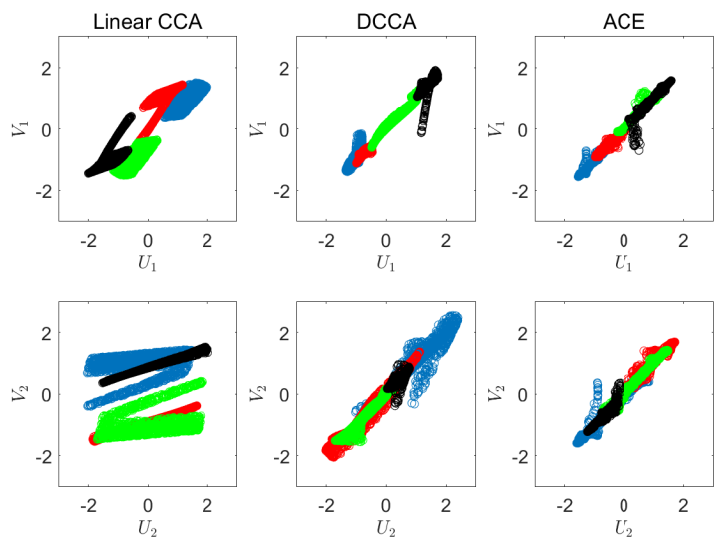}
\caption{Visualization of the correlations. Left: linear CCA. Middle: DCCA. Right:  ACE.}
\label{Circle_experiment_2}
\end{figure}

In the first experiment we visualize the outcome of our suggested CRCCA approach. Let $X$ and $Y$ be two dimensional vectors, where $X$ is uniformly distributed over the unit square and $Y$ is a one-to-one mapping of $X$, which is highly non-linear.  We draw $n=5000$ samples of $X$ and $Y$, and apply CCA, DCCA \cite{andrew2013deep}, ACE \cite{breiman1985estimating} and our suggested CRCCA method. The two charts of Figure \ref{Circle_experiment} show the samples of $X$ and $Y$, respectively. Notice that the samples' color is to visualize the mapping that we apply from $X$ to $Y$. For example, the blue samples of $X$ (which correspond to  $X_1 \in \left[0,\frac{1}{4}\right]$) are mapped to the lower left quarter circle in $Y$.  The exact description of the mapping is provided in Appendix C.  We first apply linear CCA to $X$ and $Y$. This results in a sum of correlation coefficients of $1.1$ (where the maximum is $2$). We report the normalized objective, which is  $0.55$. This poor performance is a result of the highly non-linear mapping of $X$ to $Y$, which the classical CCA attempts to recover by linear means. The two charts on the left of Figure \ref{Circle_experiment_2} visualize the correlation between the resulting components ($U_1$ against $V_1$ and $U_2$ against $V_2$). Next, we apply Deep CCA. We examine different architectures which vary in the number of layers ($3$ to $7$) and the number of neurons in each layer ($2^j$ for $j=3,\dots,12$). The remaining hyper-parameters are set according to the default values of   \cite{andrew2013deep}. We obtain a normalized objective value of $0.993$ for an architecture of three layers of $32$ neurons in each layer. This result demonstrates DCCA ability to (almost) fully recover the correlation between the original variables. The middle charts of Figure \ref{Circle_experiment_2} visualize the results we achieve for the best performing DCCA architecture.  Further, we apply the ACE method, which seeks the optimal non-parametric CCA solution to (\ref{CCA}). ACE attains a normalized objective of $0.995$ for a choice of $k=70$. The charts on the right of  Figure \ref{Circle_experiment_2} show the  correlation between the components of ACE's outcome. As expected, ACE succeeds to almost fully recover the perfect correlation of $X$ and $Y$, in this large-sample low-dimensional setup.  
 
\begin{figure}[ht]
\centering
\includegraphics[width =0.6\textwidth,bb= 120 100 670 510,clip]{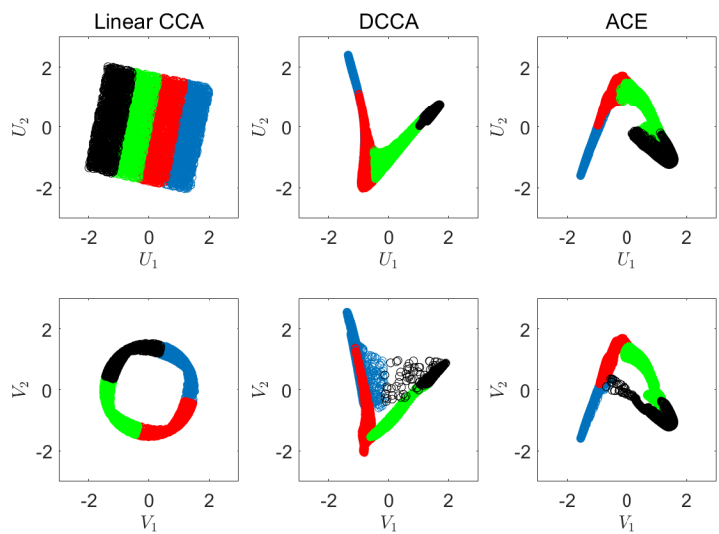}
\caption{Visualization of the obtained components. Left: linear CCA. Middle:  DCCA. Right:  ACE.}
\label{Circle_experiment_5_1}
\end{figure}

To further illustrate the applied transformations, we visualize the obtained components of each of the methods. Figure \ref{Circle_experiment_5_1} illustrates $U_1$ against $U_2$, and $V_1$ against $V_2$ for the linear CCA (left), DCCA (middle) and ACE (right). As we can see, linear CCA rotates the original vectors, while DCCA and ACE demonstrate a highly non-linear nature.

\begin{figure}[ht]
\centering
\includegraphics[width =0.6\textwidth,bb= 120 100 670 510,clip]{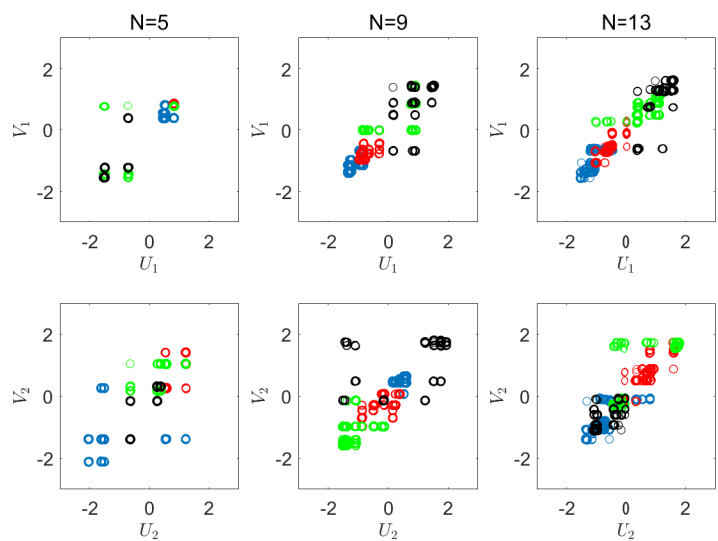}
\caption{Visualization of the correlations. CRCCA with quantization levels, $N=5,9,13$.}
\label{Circle_experiment_3}
\end{figure}

\begin{figure}[ht]
\centering
\includegraphics[width =0.6\textwidth,bb= 120 100 670 510,clip]{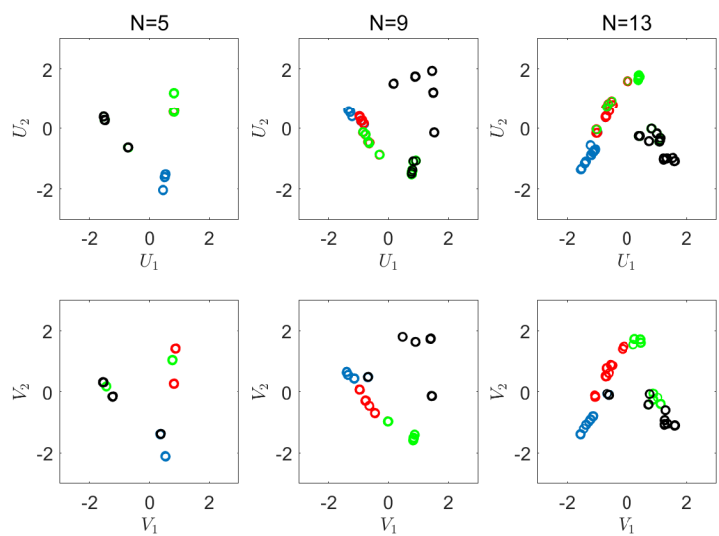}
\caption{Visualization of the obtained CRCCA components.}
\label{Circle_experiment_5_2}
\end{figure}

We now apply our suggested empirical CRCCA algorithm. Here, we use different number of quantization cells, $N=5,9,13$, where $N$ is the number of levels in each dimension. Figure \ref{Circle_experiment_3} shows the results we achieve for these three settings. The corresponding normalized objective values for each quantization level are $0.92$, $0.95$ and $0.99$, respectively. As expected, we converge to full recovery of $X$ from $Y$,  as $N$ increases. This should not come as a surprise in such an easy problem, where the dimension is low and the number of samples is large enough. As we observe the charts of Figure \ref{Circle_experiment_3}, we notice the discrete nature of CRCCA, which is an immediate consequence of the quantization we apply. As expected, we observe more diversity in the outcome of the CRCCA, as $N$ increases. In addition, we notice that the quantized points become increasingly correlated. Figure \ref{Circle_experiment_5_2} shows the obtained CRCCA components for $N=5,9,13$. Here again, we notice the discrete nature of the canonical variables, and the convergence to the optimal transformation (ACE) as $N$ increases. 


\subsection{Real-world Experiment}

Let us now turn to a real-world data experiment, in which we demonstrate the generalization abilities of the CRCCA approach. The Weight Lifting Exercises dataset \cite{velloso2013qualitative} summarizes sensory readings from four different locations of the human body, while engaging in a weight lifting exercise. Each sensor records a $13$-dimensional vector, which includes a $3$-d gyroscope, $3$-d magnetic fields, $3$-d accelerations, roll, pitch, yaw and the total acceleration absolute value. During the weight lifting activity, participants were asked to perform a specified exercise, called Unilateral Dumbbell Biceps Curl \cite{velloso2013qualitative} in five different fashions: exactly according to the specification (Class A), throwing the elbows to the front (Class B), lifting the weight only halfway (Class C), lowering the weight only halfway (Class D) and throwing the hips to the front (Class E). Class A corresponds to the specified execution of the exercise, while the other four classes correspond to common mistakes.

Since different sensors simultaneously record the same activity, we argue that they may be correlated under some transformations. To examine this hypothesis, we apply different CCA techniques to two different sensor vectors (arm and belt, in the reported experiment) to seek maximal correlations among these vectors. Specifically, we analyze an $X$ set of $13$ measurements (the arm sensor), and a $Y$ set of the same $13$ measurements for the belt sensor. Notice that we ignore the class of the activity (represented by a categorical variable from A to E) to make the problem more challenging. To have valid and meaningful results, we examine the generalization performance of our transformations. We split the dataset (of $31,300$ observations)  to $70\%$ train-set, $15\%$ evaluation-set (to tune different parameters) and $15\%$ test-set. We repeat each experiment $100$ times (that is, $100$ different splits to the three sets mentioned above) to achieve averaged results with a corresponding standard deviation. The results we report are the averaged normalized sum of canonical correlations on the test-set. 

We first apply linear CCA \cite{hotelling1936relations}. This achieves a maximal  objective value of $\bar{\rho}_{UV}=0.279(\pm0.01)$. Next, we apply kernel CCA with a Gaussian kernel \cite{lai2000kernel}. We examine a grid of Gaussian kernel variance values to achieve a maximum of $\bar{\rho}_{UV}=0.38(\pm0.09)$, for $\sigma^2=2.5$. Further, we evaluate the performance of DCCA and DCCAE. As in the previous experiment, we examine different DNN architectures, for both methods. Specifically, we look at different numbers of layers ($3$ to $7$) and different numbers of neurons in each layer ($2^j$ for $j=3,\dots,12$). In addition, we follow Andrew et al. \cite{andrew2013deep} guidelines and examine both smaller and larger mini-batch sizes, ranging from $20$ to $10000$ samples. The remaining hyper-parameters are set according to the default values, as appear in \cite{andrew2013deep} and \cite{wang2016deep}.  The best performing DCCA architecture achieves  $\bar{\rho}_{UV}=0.51(\pm0.04)$ for three layers of $2048$ neurons in each layer, and a mini-batch size of $5000$ samples. Notice that such a large mini-batch size is not customary in classical design of DNNs, but quite typical in Deep CCA architectures \cite{andrew2013deep,wang2016deep}.
DCCAE achieves  $\bar{\rho}_{UV}=0.53(\pm0.1)$ for the same architecture as DCCA in the correlation DNNs, while the autoencoders consists of three layers with $1024$ neurons in each layer, and a mini-batch size of $100$ samples.  Finally, we apply empirical ACE (via $k$-nn) with different $k$ values. We examine the evaluation-set results for  $k=10,\dots,500$ and choose the best performing $k=170$. We report a  maximal normalized objective (on the test-set) of $\bar{\rho}_{UV}=0.62(\pm0.1)$.

\begin{figure}[ht]
\centering
\includegraphics[width =0.33\textwidth,bb= 140 230 470 540,clip]{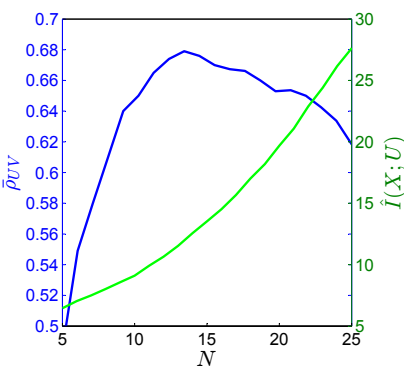}
\caption{Real-world experiment. $N$ is the number of quantization cells in each dimension. The blue curve is the objective value on the evaluation-set (left y-axis), and the green curve is the corresponding estimate of the mutual information (right y-axis).}
\label{weight_lifting_experiment_1}
\end{figure}

We now apply our suggested CRCCA method for different uniform quantization levels.  The blue curve in Figure \ref{weight_lifting_experiment_1} shows the results we achieve on the evaluation-set. The $x$-axis is the number of quantization levels per dimension, while the $y$-axis (on the left) is our objective. The best performing CRCCA  achieves a sum of correlation coefficients of $\bar{\rho}_{UV}=0.66(\pm0.1)$ for a quantization level  of $N=13$.  Importantly, we notice that the number of quantization levels determines the level of regularization in our solution and controls the generalization performance, as described in detail in Section \ref{regularization}. A small number of quantization levels implies more regularization (more values are averaged together), hence more bias and less variance. A large number of quantization levels means less regularization hence more variance and less bias. We notice that the optimum is achieved in between, for a quantization level of $13$ cells in each dimension. The green curve in  Figure \ref{weight_lifting_experiment_1} demonstrates the corresponding estimated mutual information, $I(X;U)$ (where $I(Y;V)$ is omitted from the chart as it is quite similar to it). Here, we notice that lower values of $N$ correspond to a lower mutual information while a finer quantization corresponds to a greater value of mutual information.  Notice that $U$ and $V$ are quantized versions of $X$ and $Y$ respectively, so we have that $I(X;U)=H(U)$ and $I(Y;V)=H(V)$. This demonstrates the soft dimensionality reduction interpretation of our formulation. Specifically, the green curve defines the maximal level of correlation that can be attained for a prescribed storage space (in bits). For example, we may attain up to $\bar{\rho}_{UV}=0.68$ (on the evaluation set) by representing $U$ in no more than $12.5$ bits (and a similar number of bits for $V$).

Estimating the entropy values of $U$ and $V$ is quite challenging when the number of samples is relatively small, compared to the number of cells \cite{paninski2003estimation}. One of the reasons is the typically large number of empty cells for a given set of samples. This phenomenon is highly related to the problem of estimating the missing mass (the probability of unobserved symbols) in large alphabet probability estimation. Here, we apply the Good-Turing probability estimator \cite{good1953population} to attain asymptomatically optimal estimates for the probability functions of $U$ and $V$ \cite{orlitsky2015competitive}. Then, we use these estimates as plug-in's for the desired entropy values.   

As we can see, our suggested method surpasses its competitors with a sum of correlation coefficients of $\bar{\rho}_{UV}=0.66(\pm0.1)$, for a choice of $N=13$. We notice that the advantage of CRCCA over ACE is barely statistically significant. However, on a practical note,  CRCCA takes about $20$ minutes to apply on a standard personal laptop, using a Matlab implementation, while ACE takes more than $2$ hours in the same setting. The difference is a result of the $k$-nearest neighbors search in such a high dimensional space. This search is applied repeatedly, and it is much more computationally demanding  than fixed quantization, even with enhanced search mechanisms.

To further illustrate our suggested CRCCA approach, we provide scatter-plot visualizations of the data in Appendix D. Specifically, Figures \ref{real_world_x}, \ref{real_world_y} and \ref{real_world_xy} show the arm sensor components ($X_i$ against $X_j$, for all $i,j = 1,\dots,13$), the belt sensor components ($Y_i$ against $Y_j$) and their component-wise dependencies ($X_i$ against $Y_j$), respectively. Figures \ref{real_world_u}, \ref{real_world_v} and \ref{real_world_uv} show the corresponding CRCCA components. Notice that all figures focus on the train-set samples, to improve visualization. As we can see, the arm sensor components demonstrate a more structured nature than the belt sensor components (for example, $X_1$ scatter-plots are more clustered, compared to $Y_1$ scatter-plots). As we study the different clusters, we observe that they correspond to the different weight lifting activities (class A to E, as described above). These activities focus on arm movements, which explains the more clustered nature of Figure \ref{real_world_x}, compared to Figure \ref{real_world_y}. Naturally, the structural difference between $X$ and $Y$ results in a relatively poor correlation. CRCCA reveals the maximal correlation by applying transformations to the original data. Here, we observe 
more structure in both sets of variables, which introduces a significantly greater correlation. Specifically, we notice a two-cluster structure in the first pair of canonical variates. Beyond the first pair, there is relatively less cluster structure, suggesting that the information on the two clusters is all condensed into the first canonical variate pair. It is important to mention that CCA methods are typically less effective in the presence of non-homogeneous data (for example, clustered variables as in our experiment). It is well known that the existence of groups in a dataset can provoke spurious correlations between a pair of quantitative variables. One possible solution is analyze each cluster independently (that is, apply CCA to each weightlifting activity, from A to E). However, this approach results in a smaller number of samples for each CCA, which typically decreases generalization performance.



\begin{figure}[ht]
\centering
\includegraphics[width =0.33\textwidth,bb= 250 160 540 460,clip]{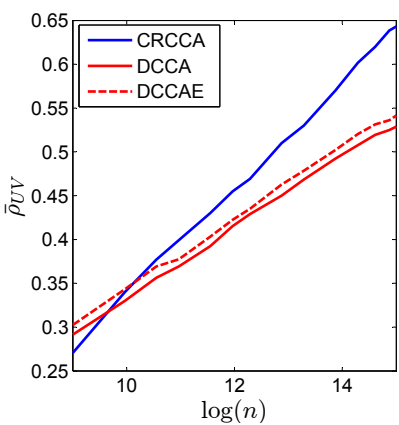}
\caption{Real-world experiment. The generalization performance of CRCCA (blue curve), DCCA (red curve) and DCCAE (dashed red) for different train-set size $n$ (in a log scale).}
\label{weight_lifting_experiment_2}
\end{figure}

To conclude, the weight lifting experiment demonstrates CRCCA ability to reveal the underlying correlation between two sets of variables; a correlation which is not apparent in the dataset's original form. In this sense CRCCA, answers our preliminary hypothesis, that the arm sensor and the belt sensor are indeed correlated. It is important to emphasize that the favorable performance of CRCCA is evident in low-dimension and large sample size setups, such as in the examples above. Unfortunately, the non-parametric nature of CRCCA makes it less effective when the dimension of the problem increases (or the number of samples reduces), due to the curse of dimensionality. Figure  \ref{weight_lifting_experiment_2} compares CRCCA with DCCA and DCCAE  for different train-set sizes. Here, we use the same DNN architectures described above. 
We notice that for relatively small sample size, CRCCA is inferior to the more robust (parametric) DCCA and DCCAE methods. However, as the number of samples grow, we observe the advantage of using CRCCA.

\section{Discussion and Conclusion}
In this work we introduce an information-theoretic compressed representation formulation of the non-linear CCA problem. Specifically, given two multivariate variables $X$ and $Y$, we consider a CCA framework in which the extracted signals $U=\phi(X)$ and $V=\psi(Y)$ are maximally correlated and, at the same time, $I(X;U)$ and $I(Y;V)$ are bounded by some predefined constants. We show that by imposing these mutual information constraints, we regularize the classical non-linear CCA problem so that $U$ and $V$ are compressed representations of the original variables. This allows us to regulate the dependencies between the mappings $\phi,\psi$ and the observations, and by that control the bias-variance trade-off and improve the generalization performance. Our CRCCA formulation draws immediate connections to the remote source coding  problem. This allows us to derive upper bounds for our generalization error, similarly to the classical rate distortion problem. In addition, we show that by imposing the mutual information constraints, we allow a soft dimensionality reduction, as opposed to the hard reduction of the traditional CCA framework. Finally, we suggest an algorithm for the empirical CRCCA problem, based on uniform quantization. 

We demonstrate the performance of our suggested algorithm in different setups. We show that CRCCA successfully recovers the underlying correlation in a synthetic low-dimensional experiment, as the number of quantization cells increases. Furthermore, we observe that the CRCCA transformations converge to the optimal ACE mapping, as we study the scatter-plots of the obtained canonical components. In addition, we apply our suggested scheme to a real-world problem. Here, CRCCA demonstrates  competitive (and even superior) generalization performance to DCCA and DCCAE at a reduced computational burden, as the number of observations increases. This makes DCCAE a favorable choice in such a regime. 

Given a joint probability distribution, our suggested CRCCA formulation  provides a sound theoretical background. However, this problem becomes more challenging in a real-world setup, where only a finite sample-size is available. In this case, our suggested algorithm drops the well-known ``conditional expectation - rate distortion" decomposition and solves the problem directly.  This results in a partition estimate, in which the cell volume is defined by the constraints of the problem. Unfortunately, just like $k$-NN, the partition estimate suffers from the curse of dimensionality. This means that the CRCCA problem may not be empirically solved in high rates, as the dimension increases. However, since the rate distortion curve is convex, one may accurately estimate the CRCCA in low rates, and interpolate the remaining points of curve. In other words, we may use the well-studied properties of the rate distortion curve in order to improve our estimation in higher rates. We consider this problem for future research. 

It is important to mention that the CRCCA is a conceptual framework and not a specific algorithm. This means that there are many possible approaches to solve (\ref{noisy_source_coding}) given a finite sample-size. In this work we focus on a direct approach using uniform lattice quantizers. Alternatively, it is possible to implement general lattice quantizers, which converge even faster to the rate-distortion bound. On the other hand, one may take a different approach and suggest any conditional expectation estimate (parametric or non-parametric), followed by a vector quantizer (as done for example by Linder et al. \cite{linder1997empirical}). Further, it is possible to derive an empirical CRCCA bound, in which we attempt to solve the CRCCA problem without quantizers at all (for example, estimate the joint distribution as a plug-in for the problem). This would allow an empirical  upper-bound for any choice of algorithm. All of these directions are subject to future research. Either way, the major contribution of our work is not in the specific algorithm suggested in Section \ref{our_suggested_method}, but the conceptual compressed representation formulation of the problem, and its information-theoretic properties.  

Finally, CRCCA may be generalized to a broader framework, in which we replace the correlation objective with mutual information maximization of the mapped signals, $I(U;V)$. This problem strives to capture more fundamental dependencies between $X$ and $Y$, as the mutual information is a statistic of the entire joint probability distribution, which holds many desirable characteristics (as shown, for example, in \cite{painsky2018universality} and \cite{painsky2019bregman}). This generalized framework may also be viewed as a two-way information bottleneck problem, as previously shown in \cite{slonim2006multivariate}.

 \authorcontributions{conceptualization, A.P., M.F. and N.T.; methodology, A.P., M.F. and N.T.; software, A.P.; validation, A.P.; formal analysis, A.P., M.F. and N.T.; investigation, A.P.; writing--original draft preparation, A.P.; writing--review and editing, M.F. and N.T.; visualization, A.P.; supervision, M.F. and N.T.; project administration, M.F. and N.T.; funding acquisition, A.P. and N.T.}

\funding{This research was supported by the Gatsby Charitable Foundation and the Intel Collaboration Research Institute for Computational Intelligence (ICRI-CI) to Naftali Tishby, and the Israeli Center of Research Excellence in Algorithms to Amichai Painsky.}

\conflictsofinterest{The authors declare no conflict of interest. The funders had no role in the design of the study; in the collection, analyses, or interpretation of data; in the writing of the manuscript, or in the decision to publish the results.} 

\appendixtitles{no} 
\appendix
\section{: A Proof for Lemma 1}

The optimal solution to (\ref{on_side_CCA}) is achieved in two steps. First, let  $\tilde{V}=\mathbb{E}(V|X)$ be the conditional expectation of $V$ given $X$. Then, $U$ is the constrained rate-distortion solution for (\ref{rate_distortion}). We may apply calculus of variations to derive the optimality conditions of (\ref{on_side_CCA}), with respect to $p(u|\tilde{v})$. Here, we present the general case where $X$,$Y$,$U$,$V$ and $\tilde{V}$ are all vectors. 
The mutual information objective is given by 
\begin{align}
I(\tilde{V};U)=&\int p(\tilde{v})p(u|\tilde{v})\log \frac{p(u|\tilde{v})}{p(u)}dud\tilde{v}.
\end{align}
while the constraints are 
\begin{enumerate}
\item $\mathbb{E}||U-\tilde{V}||^2=\int ||u-\tilde{v}||^2 p(u|\tilde{v})p(\tilde{v})dud\tilde{v}\leq D$

\item  $\mathbb{E}(U_i)=\int u_i p(u) =\int u_i p(u|\tilde{v})p(\tilde{v})dud\tilde{v}=0$

\item  $\mathbb{E}(U_i U_j)=\int u_i u_j p(u|\tilde{v})p(\tilde{v})dud\tilde{v}=\mathbbm{1}\left\{i=j\right\}$
\end{enumerate}
where $\mathbbm{1}\{\cdot\}$ is the indicator function. Therefore, our Lagrangian is given by

\begin{align}
\mathcal{L}=&\int p(\tilde{v})p(u|\tilde{v})\log \frac{p(u|\tilde{v})}{p(u)}du d\tilde{v}-\eta\left[\int ||u-\tilde{v}||^2 p(u|\tilde{v})p(\tilde{v})dud\tilde{v}-D\right]-\\\nonumber
&\sum_i \tau(i)\left[\int u_i p(u|\tilde{v})p(\tilde{v})dud\tilde{v} \right]-\sum_{i,j} \mu(i,j) \left[\int u_i u_j p(u|\tilde{v})p(\tilde{v}) -\mathbbm{1}\left\{i=j\right\} \right]-\\\nonumber
&\int \lambda(\tilde{v})\left[\int p(u|\tilde{v})du-1\right]d\tilde{v}
\end{align}
where $\eta, \tau, \mu$ are the Lagrange multipliers associated with the distortion, the mean and the correlation constraints, and $\lambda(\tilde{v})$ are the Lagrange multiplier that restrict $p(u|\tilde{v})$ to be valid distribution functions. Setting the derivative of $\mathcal{L}$ (with respect to $p(u|\tilde{v})$) to zero obtains the specified conditions.
  \hfill $\blacksquare$

\section{: A Proof for Lemma 2}

Let $Q(x)$ be a uniform quantizer of $x$ which consists of $M$ different cells. Let $C_m$ be the m$^{th}$ quantization cell. Let $u_m$ be the fit of the  m$^{th}$ cell. Specifically, $Q(x_i)=u_m$ implies that $x_i$ is a member of $C_m$. For a quantizer $Q(x)$, the empirical risk minimization of (\ref{empirical_on_side_CCA}) with respect to the fits values $u_m$, is given by 

\begin{equation} \label{lemma_opt}
\begin{aligned}
& \min_{u_m}
& & \sum_{m=1}^M \sum_{i \in C_m} ||v_i-u_m||^2 \\
& \; \text{s.t.}
& &\sum_{m=1}^M |C_m| u_m=0, \; \frac{1}{n} \sum_{m=1}^M |C_m| u_m u_m^T=I\\
\end{aligned}
\end{equation}
where $|C_m|$ denotes the number of observations in the $m^{th}$ cell. The KKT conditions of  (\ref{lemma_opt}) yield that $u_m=A\frac{1}{|C_m|}\sum_{i \in C_m} v_i +B$. However, notice that the optimal fit for the unconstrained problem is simply $\frac{1}{|c_m|}\sum_{i \in C_m} v_i$, which concludes the proof.\hfill $\blacksquare$

\section{: Synthetic Experiment Description}

Let $X$ and $Y$ be two dimensional vectors, where $X$ is uniformly distributed over a unit square and $Y$ is a one-to-one mapping of $X$, as demonstrated in Figure \ref{Circle_experiment}. Here, we describe the exact mapping from $X$ to $Y$. We begin with the blue samples.

\begin{figure}[ht]
\centering
\includegraphics[width =0.35\textwidth,bb= 60 140 750 480,clip]{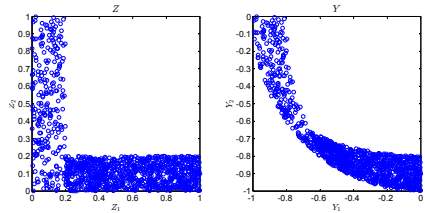}
\caption{Synthetic experiment: the blue samples of $Z_1,Z_2$ (left) and  $Y_1,Y_2$ (right).}
\label{description}
\end{figure}

First, let us spread the blue samples over the entire square. Specifically, $Z_1=4\cdot X_1$ and $Z_2=X_2$. Now, we gather all the samples to the left and lower part of the unit square, as demonstrated on the left chart of Figure \ref{description}. Specifically, $Z_1(Z_2>0.2)=0.2\cdot Z_1(Z_2>0.2)$. We shift $Z_1=Z_1-1$ and $Z_2=Z_2-1$ so that they are now located in square $[-1,0]^2$. Finally, we apply the transformation 
$Y_1=Z_1 \cdot \sqrt{1-\frac{1}{2}Z_2^2}$, $Y_2=Z_2 \cdot \sqrt{1-\frac{1}{2}Z_1^2}$
to attain a left lower quarter circle, as illustrated on the right chart of Figure \ref{description}.  Notice that the quarter circle we achieve is not homogeneous, in the sense that there are more samples for which $Z_2<0.2$ than $Z_1<0.2$. We repeat similar transformations for the rest of the samples of $X_1$ and $X_2$, so that for each quarter circle, the more dense part is clock-wise as demonstrated on the right chart of  Figure \ref{description}. 

\counterwithin{figure}{section}

\section{: Visualizations of the Real World Experiment}
\label{visualization_of_real_world}

\begin{figure}[H]
\centering
\includegraphics[width =1\textwidth,bb= 50 170 570 650,clip]{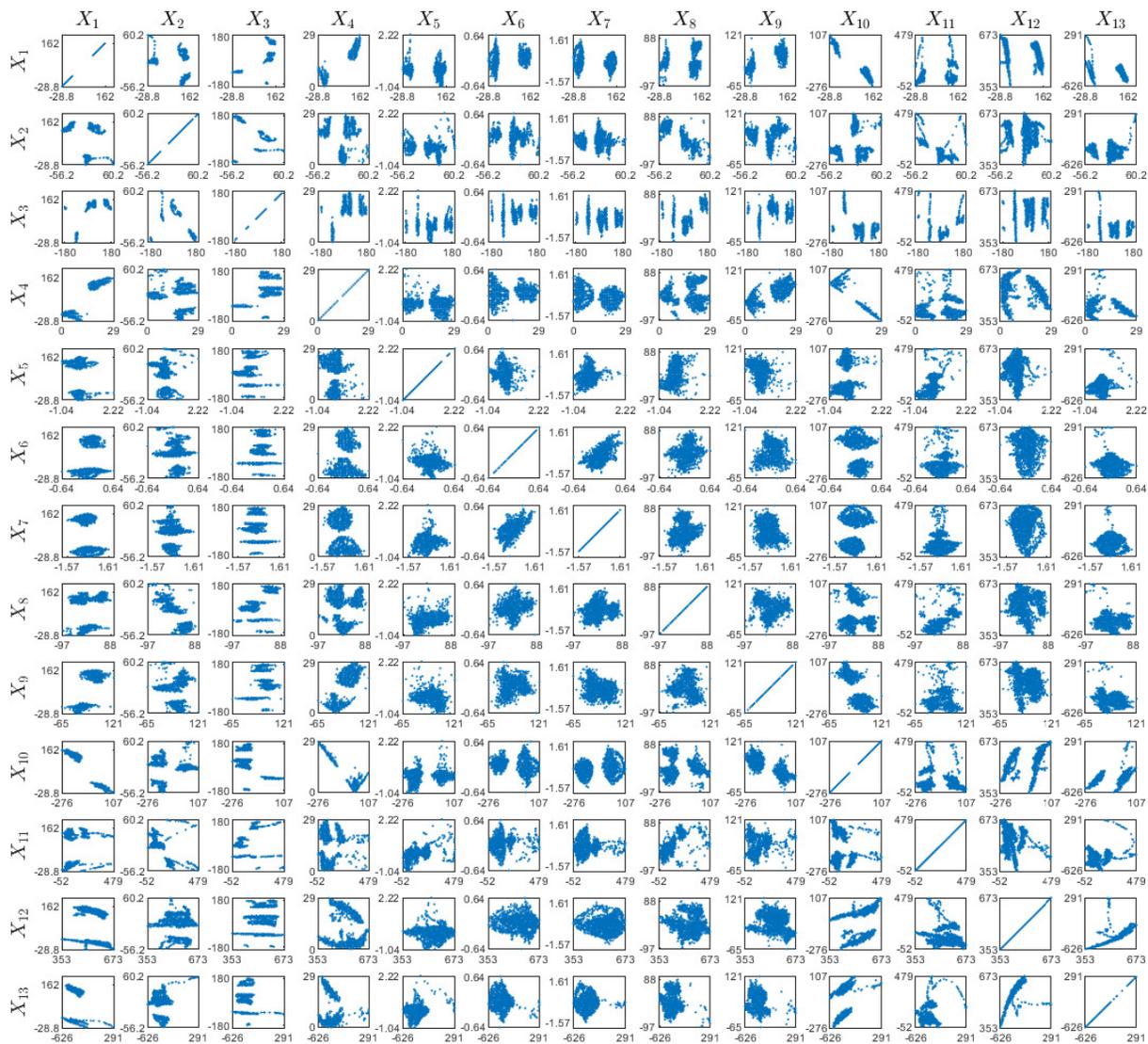}
\caption{Real-world experiment. Visualization of $X$ components (arm sensor), before CRCCA.}
\label{real_world_x}
\end{figure}

\begin{center}
\vspace*{\stretch{1}}
\begin{figure}[H]
\centering
\includegraphics[width =1\textwidth,bb= 50 170 570 650,clip]{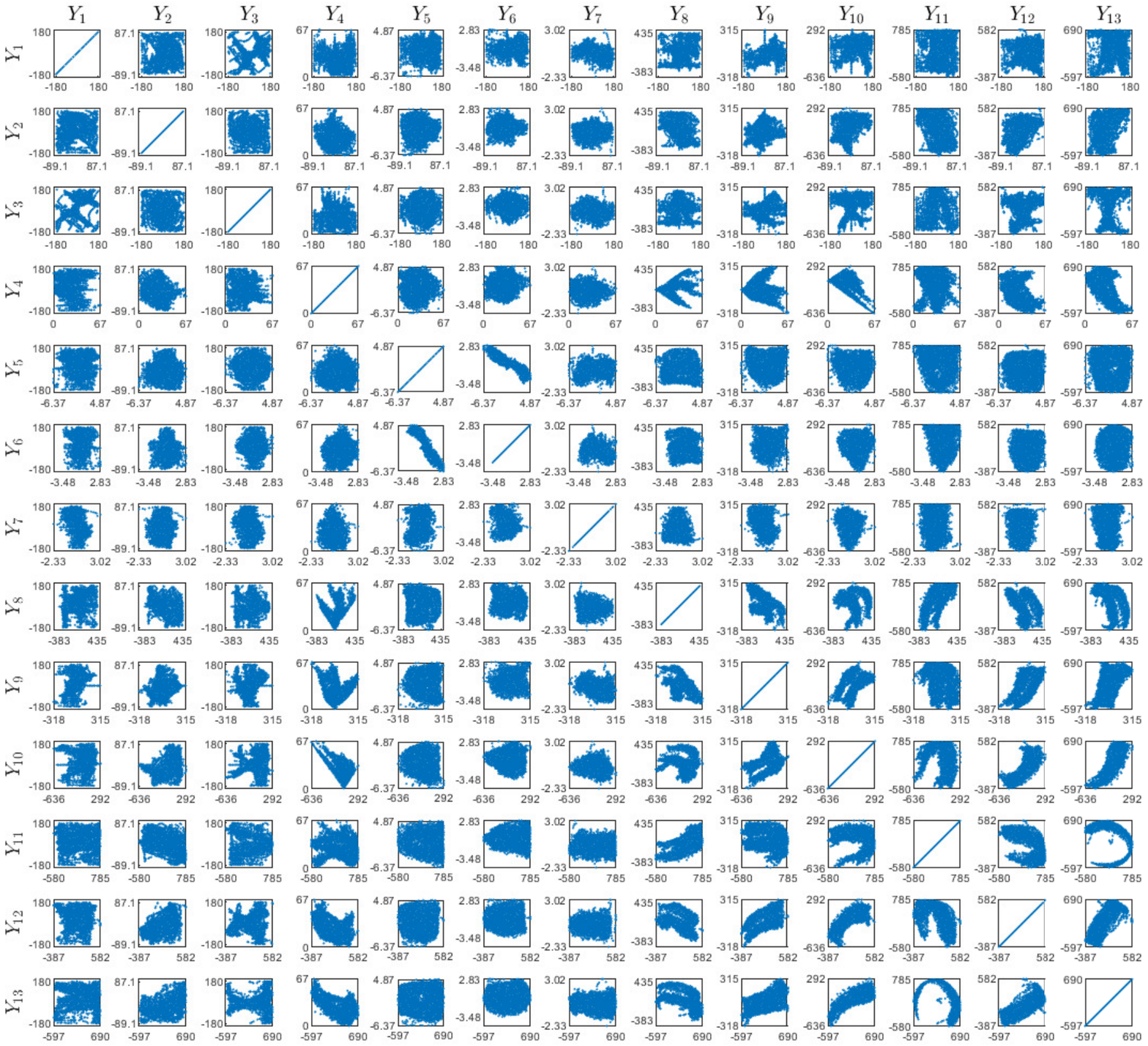}
\caption{Real-world experiment. Visualization o  $Y$ components (belt sensor), before CRCCA.}
\label{real_world_y}
\end{figure}
\vspace*{\stretch{1}}
\end{center}
\clearpage   

\begin{center}
\vspace*{\stretch{1}}
\begin{figure}[H]
\centering
\includegraphics[width =1\textwidth,bb= 50 170 570 650,clip]{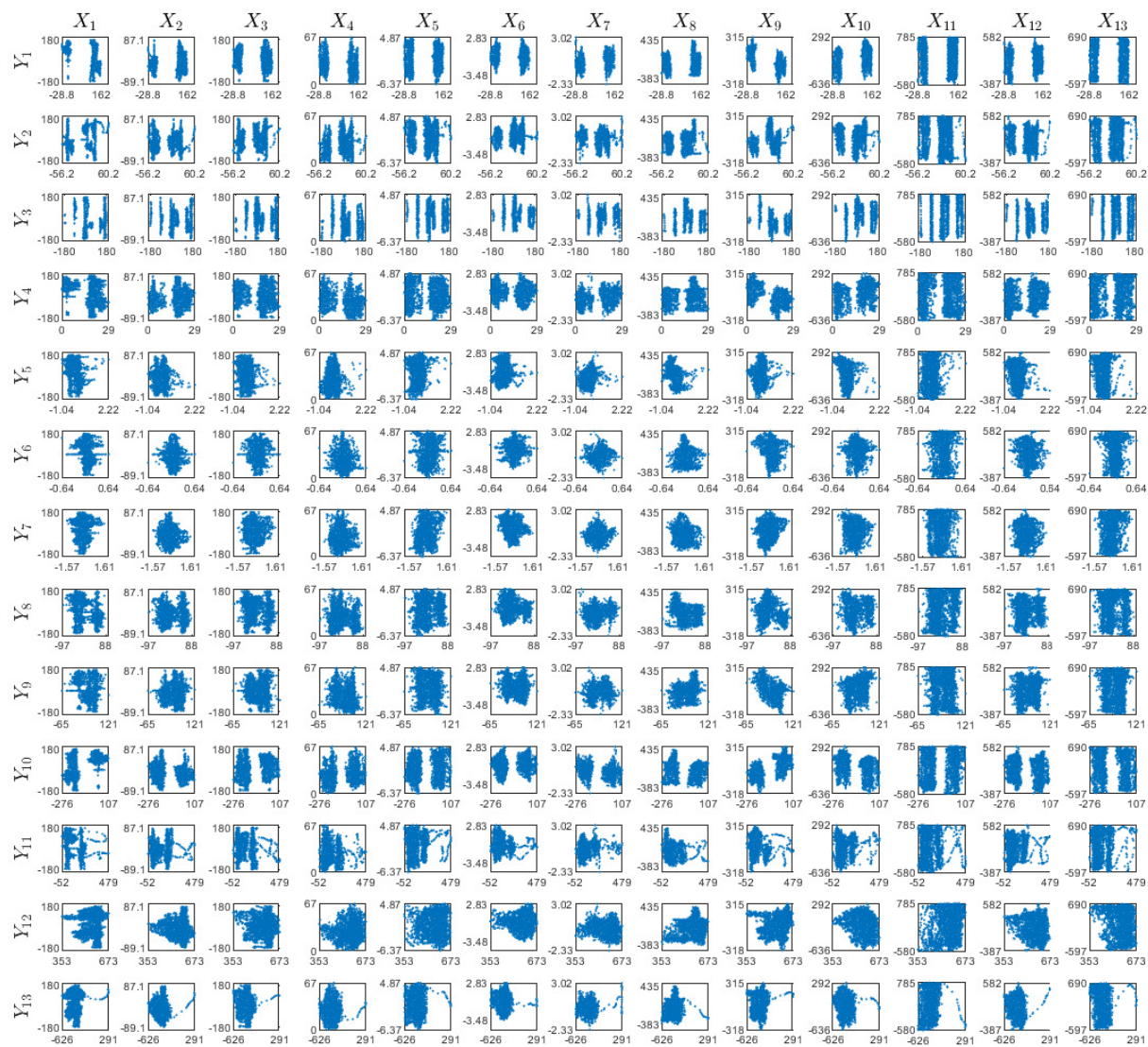}
\caption{Real-world experiment. Visualization of $X,Y$ components, before CRCCA.}
\label{real_world_xy}
\end{figure}
\vspace*{\stretch{1}}
\end{center}
\clearpage   

\begin{center}
\vspace*{\stretch{1}}
\begin{figure}[H]
\centering
\includegraphics[width =1\textwidth,bb= 50 170 570 650,clip]{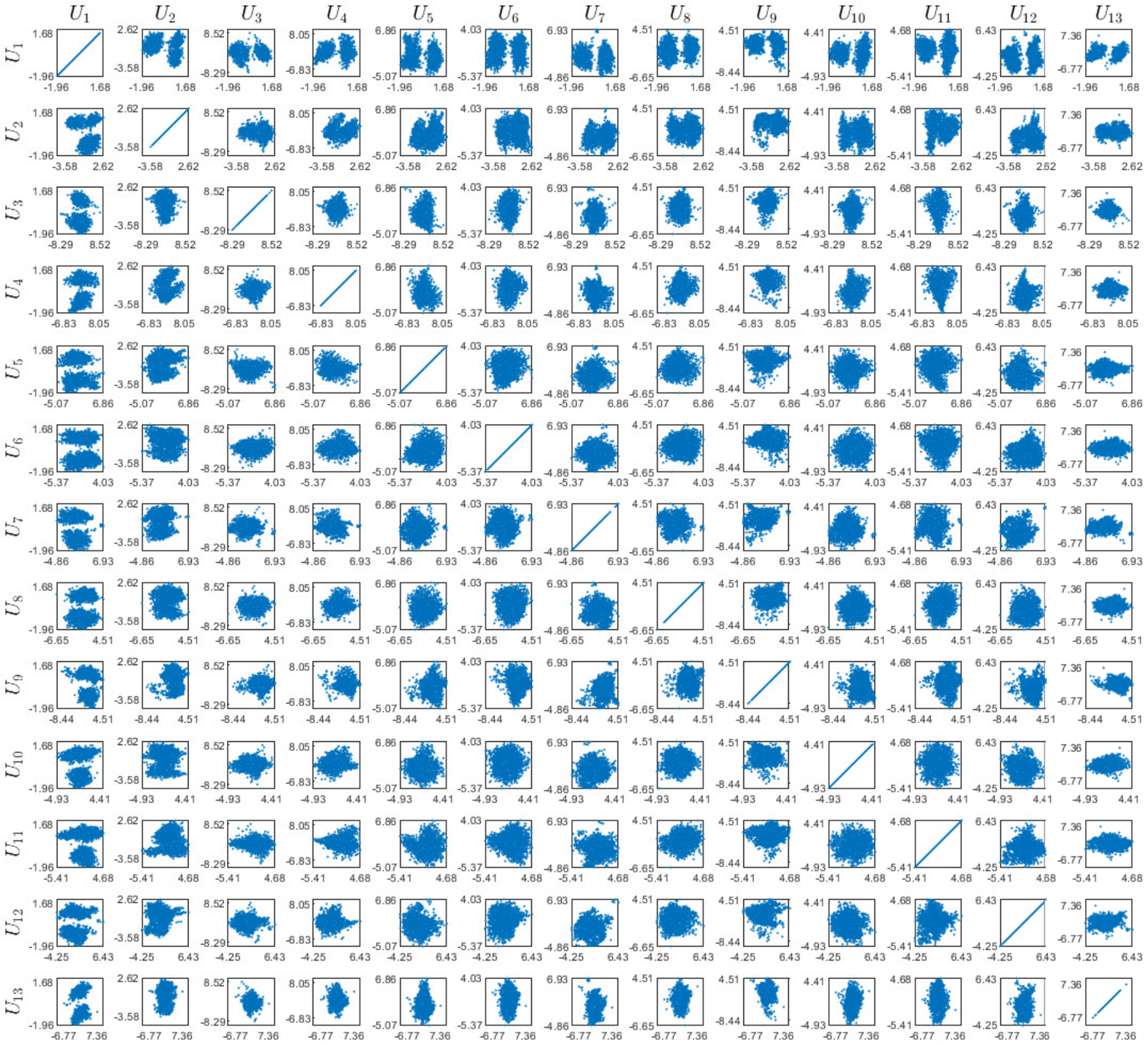}
\caption{Real-world experiment. Visualization of $U$ components, after CRCCA.}
\label{real_world_u}
\end{figure}
\vspace*{\stretch{1}}
\end{center}
\clearpage   

\begin{center}
\vspace*{\stretch{1}}
\begin{figure}[H]
\centering
\includegraphics[width =1\textwidth,bb= 50 170 570 650,clip]{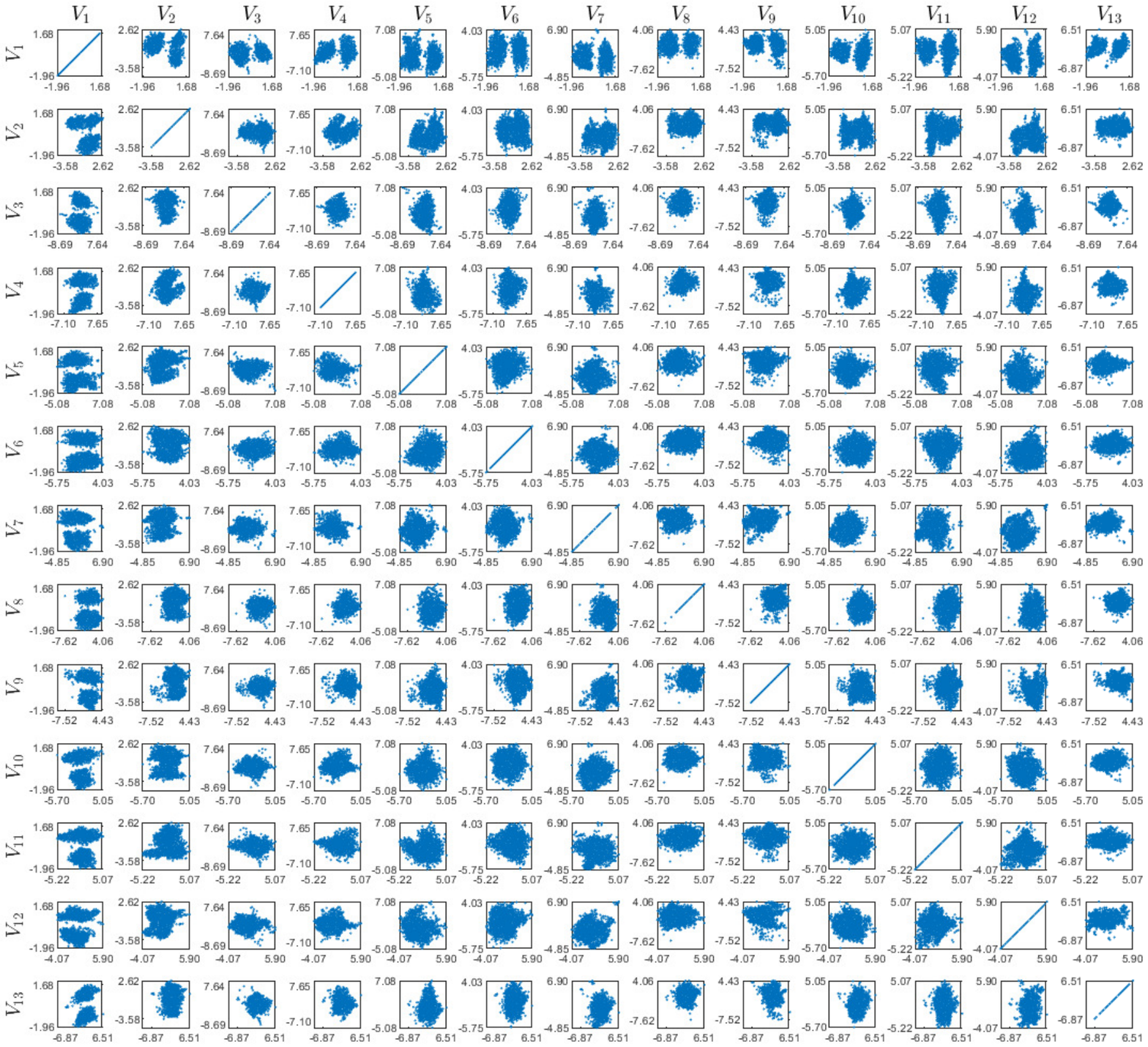}
\caption{Real-world experiment. Visualization of $V$ components, after CRCCA.}
\label{real_world_v}
\end{figure}
\vspace*{\stretch{1}}
\end{center}
\clearpage   

\begin{center}
\vspace*{\stretch{1}}
\begin{figure}[H]
\centering
\includegraphics[width =1\textwidth,bb= 50 170 570 650,clip]{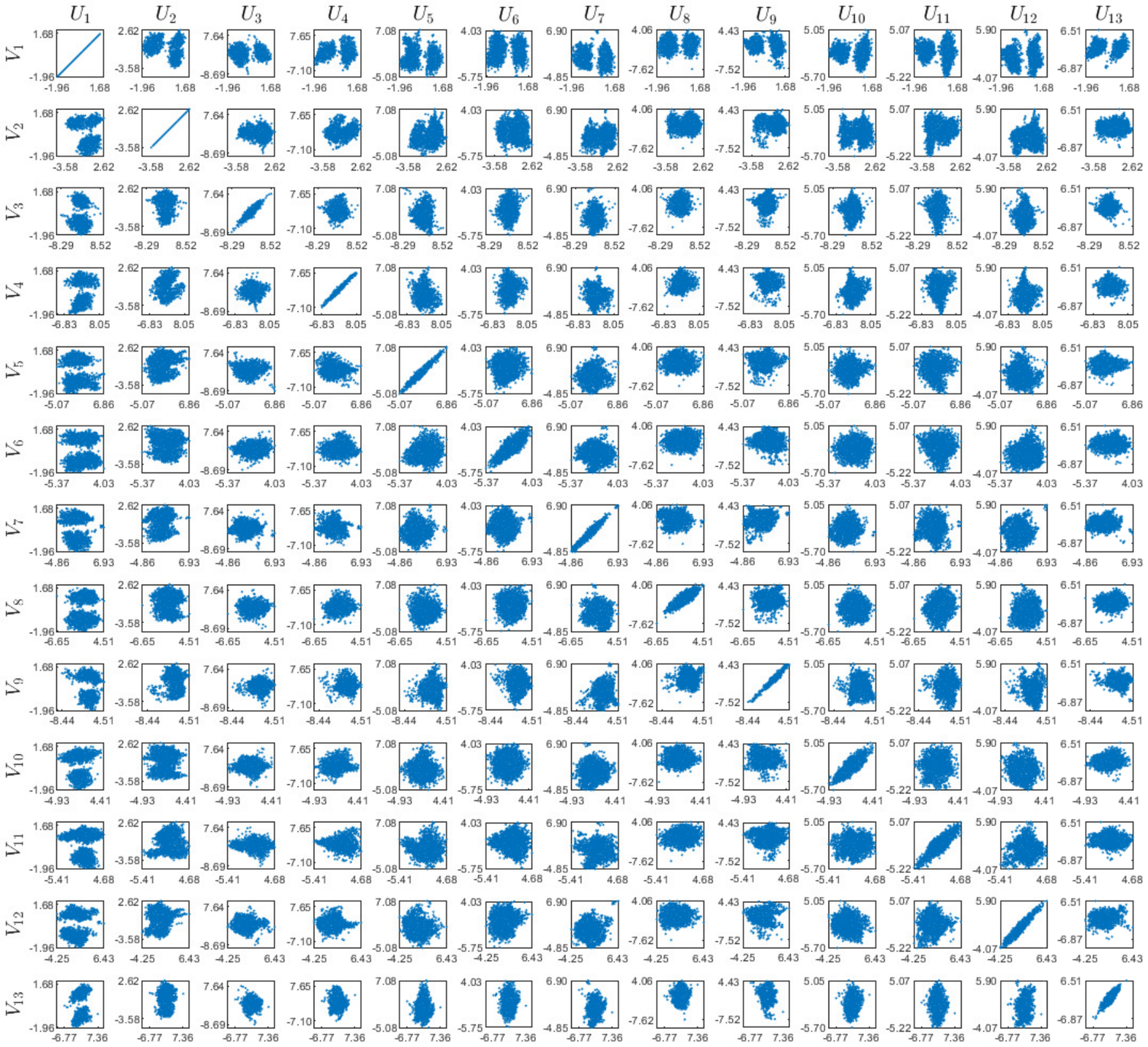}
\caption{Real-world experiment. Visualization of $U,V$ components, after CRCCA.}
\label{real_world_uv}
\end{figure}
\vspace*{\stretch{1}}
\end{center}
\clearpage

\reftitle{References}


\externalbibliography{yes}
\bibliography{sigproc}





\end{document}